\documentclass[letterpaper, 10 pt, conference]{ieeeconf}  %

\IEEEoverridecommandlockouts                              %

\usepackage{amsmath,amssymb,amsfonts}
\usepackage{graphicx}
\usepackage{textcomp}

\usepackage{verbatim}
\usepackage{tikz}

\usepackage{mathtools}
\usepackage{hyperref}
\usepackage{svg}

\DeclareMathOperator*{\argmin}{arg\,min}
\newcommand{\norm}[1]{\left\lVert#1\right\rVert}

\usepackage[backend=biber, style=ieee, doi=true, isbn=false, url=false, eprint=false]{biblatex}
\AtEveryBibitem{\clearfield{month}}
\AtEveryBibitem{\clearfield{pages}}

\newcommand{\VD}[2]{\mbox{\ensuremath{#1 \in \mathbb{R}^{#2}}}}

\newcommand{\MD}[3]{\mbox{\ensuremath{#1 \in \mathbb{R}^{#2 \times #3}}}}

\newcommand{\code}[1]{\ensuremath{\mathtt{#1}}}

\usepackage{siunitx}
\DeclareSIUnit\bar{bar}

\usepackage{algorithm, algorithmicx}
\usepackage[noend]{algpseudocode}

\newcommand*\Let[2]{\State #1 $\gets$ #2}

\newcommand{\cf}[1]{(cf. Section~#1)}

\usepackage{tabularx}

\newcommand{\diff}[0]{\mathrm{d}}

\usepackage{booktabs} %

\title{\LARGE \bf
A Soft Robotic System Automatically Learns Precise Agile Motions Without Model Information
}

\author{Simon Bachhuber\textsuperscript{1}\textsuperscript{\dag}, Alexander Pawluchin\textsuperscript{2}\textsuperscript{\dag}, Arka Pal\textsuperscript{3}, Ivo Boblan\textsuperscript{2}, Thomas Seel\textsuperscript{4}%
\thanks{\textsuperscript{\dag}Both S.B. and Al.P. contributed equally to the manuscript.}%
\thanks{\textsuperscript{1}S.B. is part of the Department Artificial Intelligence in Biomedical Engineering, FAU Erlangen-Nürnberg, 91052 Erlangen, Germany}%
\thanks{\textsuperscript{2}Al.P. and I.B. are part of the CoRoLab Berlin, University of Applied Sciences and Technology, 13353 Berlin, Germany}%
\thanks{\textsuperscript{3}Ar.P. is part of the School of Mechanical Sciences, Indian Institute of Technology Bhubaneswar, 752050 Odisha, India}%
\thanks{\textsuperscript{4}T.S. is part of the Institute of Mechatronic Systems, Leibniz Universität Hannover, 30167 Hannover, Germany}%
\thanks{Corresponding author: S.B. ({\tt\small simon.bachhuber@fau.de})}%
}

\begin{document}
	
\maketitle

\thispagestyle{empty}
\pagestyle{empty}

\begin{abstract}
    Many application domains, e.g., in medicine and manufacturing, can greatly benefit from pneumatic Soft Robots (SRs).
    However, the accurate control of SRs has remained a significant challenge to date, mainly due to their nonlinear dynamics and viscoelastic material properties.
    Conventional control design methods often rely on either complex system modeling or time-intensive manual tuning, both of which require significant amounts of human expertise and thus limit their practicality.
    In recent works, the data-driven method, Automatic Neural ODE Control (ANODEC) has been successfully used to -- fully automatically and utilizing only input-output data -- design controllers for various nonlinear systems \emph{in silico}, and without requiring prior model knowledge or extensive manual tuning.
    In this work, we successfully apply ANODEC to automatically learn to perform agile, non-repetitive reference tracking motion tasks in a real-world SR and within a finite time horizon.
    To the best of the authors' knowledge, ANODEC achieves, for the first time, performant control of a SR with hysteresis effects from only \qty{30}{\second} of input-output data and without any prior model knowledge.
    We show that for multiple, qualitatively different and even out-of-training-distribution reference signals, a single feedback controller designed by ANODEC outperforms a manually tuned PID baseline consistently.
    Overall, this contribution not only further strengthens the validity of ANODEC, but it marks an important step towards more practical, easy-to-use SRs that can automatically learn to perform agile motions from minimal experimental interaction time.
\end{abstract}

\section{Introduction}\label{sec:intro}
Soft Robots (SRs), including Pneumatic Soft Actuators (PSAs) and Pneumatic Artificial Muscles (PAMs), are gaining significant interest in diverse (bio)medical and industry application domains \cite{soft_robot_review1,habichSponge2024}.
Their rising popularity can be attributed to their inherent soft characteristics, derived from the use of flexible and deformable materials, which enables SRs to offer unique capabilities and adaptability in uncertain environments \cite{soft_robot_review2}.
These attributes prove particularly advantageous in scenarios such as rehabilitation support \cite{pneuRehaDevices} and human-robot interaction \cite{pneumaticCobot}, where providing a safe interaction environment without external sensors and safety mechanisms is crucial.

Unfortunately, the very features that give PSAs their unique advantages also introduce several challenges for their accurate control, a dilemma that is yet to be fully addressed \cite{soft_robot_review2}.
These challenges are for example: a) strong nonlinearities as the dynamics of PSAs are tightly coupled to the actuator's velocity which leads to lower control accuracy and overshoot, and b) creeping and hysteresis (cf.~Figure~\ref{fig:training_data}, bottom subplot) due to the viscoelasticity and time-dependent properties of the soft material.

\begin{figure}[t]
    \centering
    \begin{tikzpicture}
    \node[anchor=north west,inner sep=0] at (0,0) {\includegraphics[scale=1]{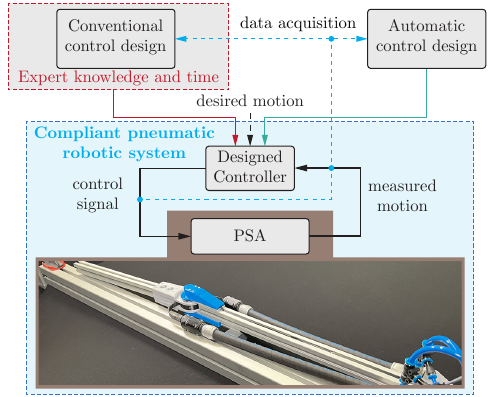}};
    \end{tikzpicture}
	\caption{
Comparison between conventional controller design, reliant on human expertise, and an automatic, data-driven control structure for Pneumatic Soft Actuators (PSAs). Both controller design paradigms observe the PSA's control signal and resulting measured motion, and subsequently design position feedback controllers that enable the measured motion of the PSA to track the desired motion. In this work we show that the data-driven method, Automatic Neural ODE Control (ANODEC), enables automatic control design for a PSA.}
\label{fig:figure1}
\end{figure}

Despite these challenges, several methods have been previously applied for the control of PSAs.
Model-based methods have been applied for PSAs such as, e.g., \cite{soft_robot_model_based_control, soroControl_survey, Falkenhahn2017, Martens2018} but they typically require a detailed system model for accurate control and therefore involve extensive system modeling combined with human expertise.

Data-driven methods can alleviate the need for extensive system modeling and have been applied for PSAs.
In \cite{Haggerty2023,soroKoopmanControl}, Koopman operators are used to achieve reference tracking of a SR, however, they make restrictive assumptions such as either weak hysteresis characteristics or non-agile reference trajectory targets.
In \cite{Gillespie2018, Johnson2021, Hyatt2019}, a DNN-based MPC was employed to control SRs but they typically focus on non-agile motions and within a computationally intensive framework.
In \cite{kasaei2023a}, an ODE-based kinematic model is learned and combined with MPC in order to learn to perform non-agile, quasistatic motions.
In \cite{Hofer2019,Schndele2012,MaBucSchMue22}, Iterative Learning Control (ILC) in combination with different feedback controllers are used for the control of PSAs.
However, ILC is tailored to repetitive motions and requires a series of iterations.
Only in \cite{MaBucSchMue22}, deep learning is used to interpolate from previously learned trajectories and to achieve a broader range of motions, but they require large amounts of system interaction time to learn a single motion.

Recently, the data-driven method, Automatic Neural ODE Control (ANODEC) has been shown to -- fully automatically -- design feedback controllers for various nonlinear systems \emph{in silico}, and without requiring prior model knowledge or extensive manual tuning.
Unlike various other data-driven methods, e.g., \cite{deisenrothPILCOModelBasedDataEfficient2011, schulmanProximalPolicyOptimization2017, bevandaDatadrivenLQRKoopmanizing2022, dataDrivenMPC, bristowSurveyIterativeLearning2006, meindlBridgingReinforcementLearning2021}, ANODEC requires neither the full state to be observed nor does it require an approximate system model.

In this contribution, we successfully apply ANODEC to automatically learn to perform agile, non-repetitive reference tracking motion tasks in a real-world PSA within a finite time horizon.
To the best of the authors' knowledge, ANODEC achieves, for the first time, performant control of a PSA with hysteresis effects (cf.~Figure~\ref{fig:training_data}) from only \qty{30}{\second} of input-output data and without any prior model knowledge.
Additionally, the single feedback controller designed by ANODEC enables tracking of arbitrary reference signals and is validated on multiple, qualitatively different, and even out-of-training-distribution reference signals.
Moreover, it is shown that ANODEC outperforms a manually tuned PID baseline, which similar to ANODEC, requires no prior model knowledge and does not require the system state to be observed.

Overall, it is shown that ANODEC allows for automatic control design (cf.~Figure~\ref{fig:figure1}) and can enable PSAs to learn to perform agile motions from only a parsimonious amount of experimental interaction time and without requiring any prior model knowledge.

\section{Pneumatic Soft Actuator}\label{sec:problem_formulation}
\begin{figure}[t]
    \centering
    \begin{tikzpicture}
    \node[anchor=north west,inner sep=0] at (0,0) {\includegraphics[scale=1]{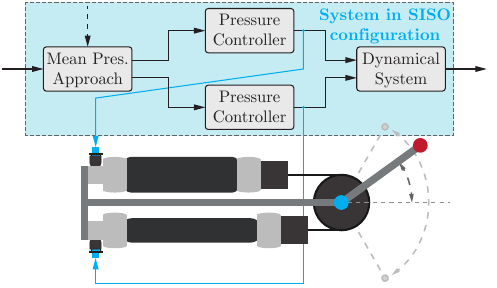}};
    \draw (2mm, -10mm) node {\small $u$};
    \draw (12mm, -5mm) node {\small $p_\text{m}$};
    \draw (32mm, -3mm) node {\small $p_\text{d,1}$};
    \draw (32mm, -16mm) node {\small $p_\text{d,2}$};
    \draw (53mm, -8mm) node {\small $p_\text{1}$};
    \draw (53mm, -15.5mm) node {\small $p_\text{2}$};
    \draw (79mm, -10mm) node {\small $\varphi$};
    \draw (67mm, -32mm) node {\small $\varphi$};
    \draw (12mm, -26mm) node {\small $p_\text{1}$};
    \draw (54.5mm, -26.5mm) node {\small $F_\text{1}$};
    \draw (12mm, -42.5mm) node {\small $p_\text{2}$};
    \draw (54.5mm, -42.5mm) node {\small $F_\text{2}$};
    \draw (60mm, -21mm) node {\small $\varphi_\text{max}$};
    \draw (60mm, -47mm) node {\small $\varphi_\text{min}$};
    \end{tikzpicture}
    \caption{The PSA including two PAMs is controlled using two pressure controllers, which implement a SISO control strategy through the mean and difference pressure approach. As a result, the single control input $u$ controls both desired difference pressures $p_\text{d,1/2}$, and the single system output is the hinge joint angle $\varphi$.}
	\label{fig:figure2}
\end{figure}
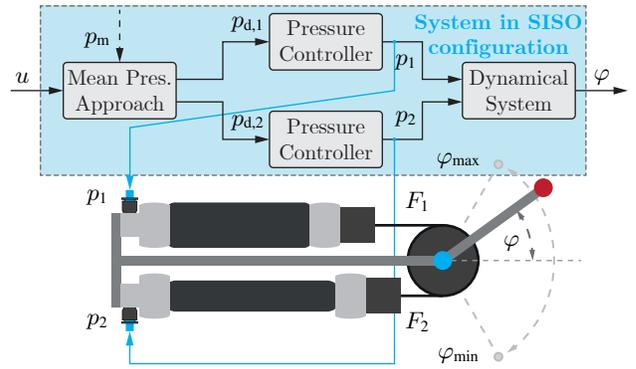

We consider a PSA, as shown in Fig.~\ref{fig:figure2}, that typically consists of a rigid or continuous kinematic structure and is driven by an antagonistic pair of PAMs. The muscles are attached via a belt and form a 1-Degree-of-Freedom (DOF) arm. The hinge joint angle represents the system output $\varphi \in \mathbb{R}$ and it is limited by the geometrical configuration of the muscle contraction and diameter of the pulley such that the range of motion $\varphi \in [\varphi_\text{min}, \varphi_\text{max}]$ where $\varphi_\text{min/max} \in \mathbb{R}$.

Each of the PAMs exhibits undesired parasitic hysteresis and creep properties that strongly influence the nonlinear characteristic force map $F_\text{1,2} \in \mathbb{R}$.
Given the unknown forces $F_\text{1}$ and $F_\text{2}$, we instead control the pressure in the bellows.
For this purpose, we use two model-based pressure controllers, which are designed to follow the given desired pressure $p_\text{d,1} \in \mathbb{R}$ and $p_\text{d,2} \in \mathbb{R}$.

Based on this system description, the system output is given by
\begin{equation}
	\varphi(t) = \tilde{\Psi}\left(p_\text{d,1}(t' < t), p_\text{d,2}(t' < t)\right) \quad \forall t \in [0, T], \label{eq:sys_miso}
\end{equation}
where $\tilde{\Psi}$ captures the system's dynamics (including the two pressure controllers), $T \in \mathbb{R}$ is the finite trial duration and where $p_\text{d,1/2}(t' < t)$ is used to denote that $\varphi(t)$ depends on all desired pressures up to time $t$, i.e., $p_\text{d,1/2}(t') \; \forall t'<t$. 

Since the PSA has two pressure inputs, we couple them using the mean pressure approach in~\eqref{eq:u} to reduce the input dimension which then yields a Single-Input Single-Output (SISO) system.
This is done by defining a mean pressure $p_\text{m} = \text{const.} \in \mathbb{R}$ and a difference pressure $\Delta p_\text{d} = p_\text{d,1} - p_\text{d,1}$, i.e.,
\begin{equation}
    p_\text{d,1} = p_\text{m} + 0.5 u \quad \text{and} \quad p_\text{d,2} = p_\text{m} - 0.5 u.
    \label{eq:u}
\end{equation}
The difference pressure becomes the system's single control input $u \in \mathbb{R}$, i.e., $u := \Delta p_\text{d}$, which is limited by the maximum physical permissible pressure range $u_\text{min} \in \mathbb{R}$ and $u_\text{max} \in \mathbb{R}$. 
Finally, using \eqref{eq:sys_miso} and \eqref{eq:u}, the system in SISO configuration is given by
\begin{equation}
	\varphi(t) = \Psi\left(u(t' < t)\right) \quad \forall t \in [0, T]. \label{eq:sys}
\end{equation}
where the functional $\Psi$ is called the system input-output map.

In this work, we aim to design a feedback controller that manipulates~$u(t)$ to let $\varphi(t)$ follow a given time-varying reference signal~$\varphi_\text{d}(t) \in \mathbb{R}$. Thus, we seek to find a controller $\Phi$ that maps the desired angle $\varphi_\text{d}(t)$ and the measured output $\varphi(t)$ to the input vector $u(t)$, that is,
\begin{equation}
    u(t) = \Phi\left(\varphi_\text{d}(t' < t), \varphi(t' < t)\right) \quad \forall t \in [0, T], \label{eq:ctrb}
\end{equation}
such that it minimizes the tracking error between the output and the reference signal, i.e.,
\begin{equation}
    \Phi^* = \argmin_\Phi \int_0^T \norm{\varphi_\text{d}(t) - \varphi(t)}_2 \diff t
\end{equation}
over the finite trial duration $T$.
Note that the desired feedback controller measures the system output and not the state and that an arbitrary reference signal can be provided in real time and must not be provided ahead of time.
However, we assume that the system captured by $\Psi$ is repeatedly reset to some trial-invariant initial state before each trial, and that the open-loop system can be safely probed for input-output data.
Note that the latter assumption is not only true for open-loop stable systems but can also be true for unstable system where the instability acts on large timescales relative to the finite trial duration, i.e., it is sufficiently slow.

\section{Automatic Neural ODE Control (ANODEC)}\label{sec:anodec}
Internally, the ANODEC algorithm \cite{anodec} designs controllers in a two-step approach of model learning from input-output data pairs, and controller learning that utilizes the learned model.

\begin{figure}
    \hspace{0.01\textwidth}
    \includegraphics[width=0.4\textwidth]{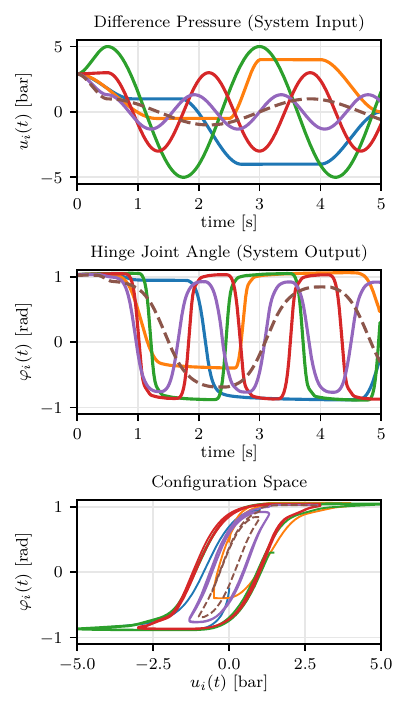}
    \vspace{-2mm}
    \caption{Training data for ANODEC consists of five input-output pairs, each of a five seconds length, that are gathered from the experimental PSA. One additional input-output pair (dashed line) is collected and used as validation data and to prevent model overfitting. The feasible input and output intervals of the experimental PSA are $u(t) \in [-6, 6] \, \qty{}{\bar}$ and $\varphi(t) \in [-1, 1] \, \qty{}{\radian}$, respectively.}
    \label{fig:training_data}
\end{figure}

\subsection{Data Requirements}\label{sec:sub_data_requirements}
To apply ANODEC, a dataset $\mathcal{D}$ that comprises of six pairs of input-output data over time, each corresponding to one trial of length $T=\qty{5}{\second}$, is gathered from the system input-output map $\Psi$.
The data is recorded at \qty{100}{\hertz}.
Fig.~\ref{fig:training_data} shows the six input-output data pairs.

The model learning~\cf{\ref{sec:sub_model_learning}} then uses the dataset
\begin{equation}\label{eq:dataset}
	\mathcal{D} = \{(u_i(t), \varphi_i(t)) | i \in 1\dots 6\}.
\end{equation}
The sixth pair of input-output data is split from the dataset~$\mathcal{D}$ and used as validation data.
The validation data is used to ensure that the learned model is accurate and that the amount of collected data is sufficient for an accurate model.
Thus, in total ANODEC requires \qty{30}{\second} of interaction time with the PSA.
Input trajectories $u(t)$ for gathering the data and thus probing the system are drawn from a sinusoidal and a spline function generator.
These functions are provided as pseudocode in Algorithm~\ref{alg:draw_u}.
Before each trial starts, the system enters a saturation state such that $\varphi(t=0) \approx \varphi_\text{max}$. This procedure is required as otherwise there might be state ambiguity due to the hysteresis of the PAMs.
Note that the system probing and model learning step \cf{\ref{sec:sub_model_learning}} can easily be extended to an iterative process that switches between system probing and model learning until a sufficiently accurate model is obtained.
This way ANODEC can be extended to a fully autonomous control design algorithm that iteratively improves the performance of the designed feedback controller.

\subsection{Model Learning}\label{sec:sub_model_learning}
ANODEC approximates $\Psi$ in \eqref{eq:sys} by a neural ODE that is learned from pairs of experimental input-output trajectories $u_i(t), \varphi_i(t)$ \cf{\ref{sec:sub_data_requirements}}.
The neural ODE that approximates~$\Psi$ is given by
\begin{align}
    \frac{\diff \xi^{(m)} (t)}{\diff t} &= \tanh\left(A_1^{(m)} \left(\xi^{(m) \intercal}(t),\, u(t)\right)^\intercal
    + b_1^{(m)} \right), \label{eq:model_ode}\\
    \hat{\varphi}(t) &= A_2^{(m)} \xi^{(m)}(t) + b_2^{(m)}, \nonumber
\end{align}
where the latent state vector \VD{\xi^{(m)}}{9}, and with model parameters \MD{A_1^{(m)}}{9}{10}, \VD{b_1^{(m)}}{9}, \MD{A_2^{(m)}}{1}{9}, $b_2^{(m)} \in \mathbb{R}$. We denote the vector of all model parameters by $\theta^{(m)} \in \mathbb{R}^{109}$ and use supervised learning to estimate and then optimize for these parameters
\begin{equation}
    \theta^{(m)^*} = \argmin_{\theta^{(m)}} \sum_{i=1}^5
    \int_0^T \norm{\varphi_i(t) - \hat{\varphi}_i(t)}_2 \diff t.
    \label{eq:model_learning}
\end{equation}
\vspace{1mm}

\subsection{Controller Learning}\label{sec:sub_controller_learning}
As a second step, ANODEC designs a feedback controller using the trained model with frozen parameters $\theta^{(m)^*}$. 
Internally, it performs a very large number of forward simulations of the closed-loop dynamics (of trained model and feedback controller) with randomly drawn step reference signals.

ANODEC represents the controller $\Phi$ as a neural ODE, given by
\begin{align}
    \frac{\diff \xi^{(c)} (t)}{\diff t} &= A_1^{(c)} \left(\xi^{(c) \intercal}(t),\, \varphi(t),\, \varphi_\text{d}(t)\right)^\intercal + b_1^{(c)}, \label{eq:controller_ode}\\
    \bar{u}(t) &= \tanh\left(A_2^{(c)} \xi^{(c)}(t) + b_2^{(c)}\right), \nonumber\\
    u(t) &= \left(u_\text{max} - u_\text{min}\right)\left(\bar{u}(t)\,0.5 + 0.5\right) + u_\text{min}, \nonumber
\end{align}
where the latent state vector \VD{\xi^{(c)}}{5}, and with controller parameters \MD{A_1^{(c)}}{5}{7}, \VD{b_1^{(c)}}{5}, \MD{A_2^{(c)}}{1}{5}, $b_2^{(c)} \in \mathbb{R}$. We denote the vector of controller parameters by $\theta^{(c)} \in \mathbb{R}^{46}$.

ANODEC then closes the loop between the neural ODEs that approximate the system input-output map $\Psi$ \eqref{eq:model_ode} and the controller $\Phi$ \eqref{eq:controller_ode}. 
The resulting closed-loop ODE takes the reference $\varphi_\text{d}(t)$ as an input and outputs $\hat{\varphi}(t)$, and we optimize the controller parameters $\theta^{(c)}$ via
\begin{multline}
    \theta^{(c)^*} = \argmin_{\theta^{(c)}} \Bigg( \lambda^{(c)} \norm{\theta^{(c)}}_2 + \\
    \underset{\varphi_\text{d} \sim \mathcal{U}(\varphi_\text{min}, \varphi_\text{max})}{\mathbb{E}}
    \left[ \int_0^T \norm{\varphi_\text{d} - \hat{\varphi}(t)}_2 \diff t \right] \Bigg)
    \label{eq:controller_learning}
\end{multline}
where $\lambda^{(c)} = 4 \cdot 10^{-4}$ is a small regularization weight, and the reference $\varphi_\text{d}$ is a constant value drawn uniformly from $[\varphi_\text{min}, \varphi_\text{max}]$.

\subsection{Time Integration \& Optimization}
We use Runge-Kutta of fourth order (RK4) to obtain a numerical solution of the time integration in \eqref{eq:model_learning} and \eqref{eq:controller_learning}, together with the initial condition $\xi^{(m)}(t=0) = 0$ and $\xi^{(m)}(t=0) = \xi^{(c)}(t=0) = 0$, respectively.

For model training, the five input-output pairs from the training dataset $\mathcal{D}$ are combined into a single model training batch. 
For controller training, the expectation operator $\underset{\varphi_\text{d} \sim \mathcal{U}(\varphi_\text{min}, \varphi_\text{max})}{\mathbb{E}}$ is estimated by randomly drawing \qty{50}{} constant reference values at every optimizer step.

Both \eqref{eq:model_learning} and \eqref{eq:controller_learning} are optimized using the Adam optimizer with a learning rate of \qty{1e-3}{} and global gradient L2 norm clipping of \qty{1.0}{}. Gradients are computed using backpropagation.
The model is trained for \num{50000} optimizer steps and the single validation input-output pair is used for early stopping and thus prevent overfitting during the model learning. The controller is trained for \num{12000} optimizer steps.

\begin{figure}
    \centering
    \begin{tikzpicture}
    \node[anchor=north west,inner sep=0] at (0,0) {\includegraphics[width=.48\textwidth]{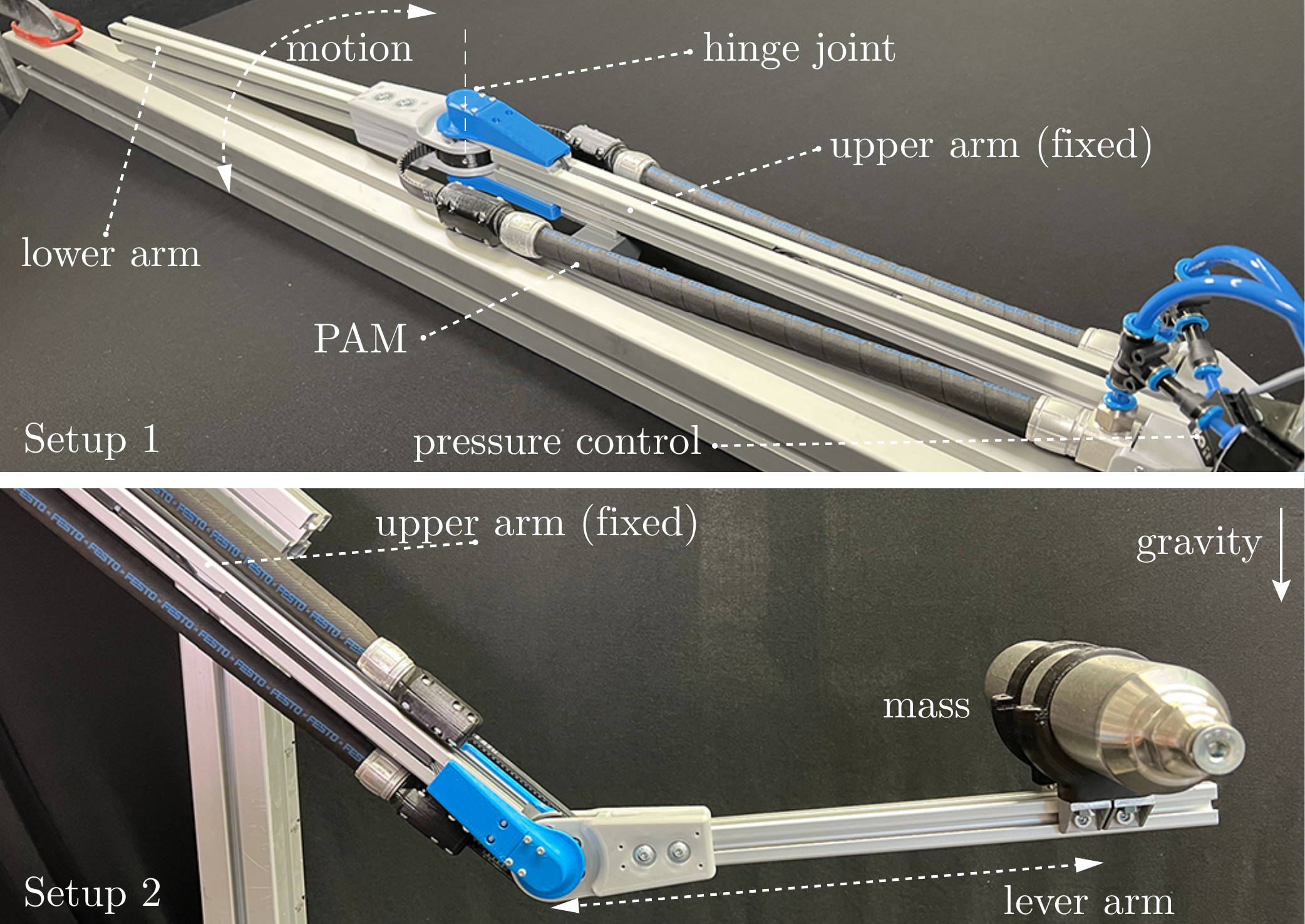}};
    \draw (29mm, -3.8mm) node {\small \color{white}$\varphi$};   
    \end{tikzpicture}
    \vspace{-3mm}
    \caption{Two experimental setups of a pneumatic arm with a single DOF. In both setups, two PAMs (pneumatic artificial muscles, black tubes) are used as an antagonistic pair to control the arm's forces and position. The upper setup shows the simplest configuration without the influence of gravity and external load. The lower setup is loaded with an external weight of \SI{0.6}{kg}, with a lever arm of \SI{0.25}{m} oriented against gravity. Both arms are fixed to the ground to prevent undesired movements.}
	\label{fig:pneumatic_system}
\end{figure}

\begin{figure*}
    \centering
    \includegraphics[width=0.9\textwidth]{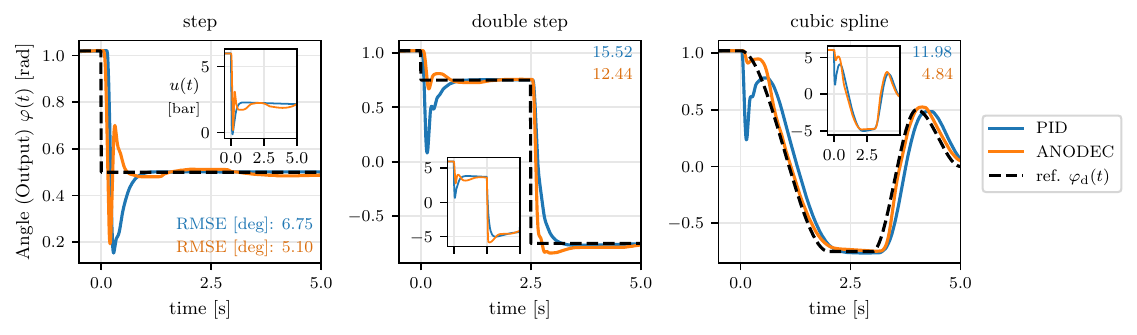}
    \vspace{-3mm}
    \caption{Performance comparison of the automatic ANODEC and a manually tuned PID controller baseline in Setup 1 for one exemplary reference signals drawn from the three reference signal distributions: Step, double step, and smooth. Even on reference signals that were not present during training (double step, smooth), ANODEC is able to consistently outperform the PID baseline and achieves a lower RMSE tracking error while requiring less experimental interaction time. Video (\href{https://youtu.be/7HkXKy0WuRw}{\scriptsize \texttt{https://youtu.be/7HkXKy0WuRw}}) that showcases these trials.}
    \label{fig:main_results}
\end{figure*}

\section{Experimental Validation}\label{sec:results}
In this section, we validate ANODEC~\cf{\ref{sec:anodec}} on an experimental PSA~\cf{\ref{sec:problem_formulation}}.
To showcase the generalization capabilities of ANODEC in Section~\ref{sec:generalization_on_unseen_reference_signals}, we propose unseen reference signal distributions that, in Section~\ref{sec:sub_results}, will be used to validate ANODEC's performance compared to a PID controller baseline~\cf{\ref{sec:sub_pid_baseline}}. The corresponding demonstration video to the presented results in Fig.~\ref{fig:main_results} can be found \href{https://youtu.be/7HkXKy0WuRw}{\underline{here}}.

\subsection{Experimental setup}\label{sec:exp_setup}
We evaluate ANODEC's performance on a pneumatic compliant robotic arm which should solve a pick and place motion task on a given reference trajectory. The arm, as depicted in Fig.~\ref{fig:pneumatic_system}, consists of both an upper and a lower arm. The upper arm carries the PAMs responsible for controlling the motion of the lower arm. The hinge joint has a range of motion within $\varphi \in [\qty{-1}{}, \qty{1}{}] \, \qty{}{\radian}$ and connects both PAMs (DMSP-10-250N) via a constant pulley. The angle is measured using a MLX90316 rotary position sensor. The upper setup (Setup 1) illustrates the configuration without external influences such as gravity and load. The lower setup (Setup 2) shows the arm loaded with a \SI{0.6}{kg} weight on a \SI{0.25}{m} lever arm against gravity. Due to the non-negligible mass of the lower arm and the highly agile target motions, the upper arm is securely fixed to the ground to prevent undesired movements.

To generate forces, the PAMs are driven using a model-based pressure controllers and two proportional directional control valves MPYE-5-M5-010-B for controlling the mass flow. The controlled pressure is measured using two SPTE-P10R-S4-B-2.5K pressure transmitters. The supply pressure was set to \qty{8}{bar}. This results in a feasible system input range $u \in [\qty{-6}{}, \qty{6}{}] \, \qty{}{\bar}$.

The pressure controllers runs on an embedded computer with a sampling rate of \qty{200}{Hz}. This system is connected through Ethernet and ROS to a computer (Intel i7 7700) that controls both ANODEC and PID approaches in Python with a control rate of \qty{100}{Hz}.

\subsection{Baseline Controller}\label{sec:sub_pid_baseline}
As a baseline, we compare the performance of \mbox{ANODEC} to a manually tuned PID controller, which similar to \mbox{ANODEC}, requires no prior model knowledge and does not require the system state to be observed.
Eleven trials were required (equaling \qty{55}{\second} of interaction time) for manual tuning and to achieve a sufficiently performant PID controller.
Since we only control the pressure and have no assumptions regarding the force-pressure relation, an integral part is needed to compensate for the PAMs force nonlinearity and hysteresis.

The tuned gain parameters are $k_\text{p} = 2, k_\text{i} = 30$ and since the derivative term of the PID controller is highly sensitive to the sensor noise from the angle sensor, leading rapidly to undesired instabilities, the D term is set to zero. The PID controller's output $u$ was clipped to stay within the feasible system input range of $[u_\text{min}, u_\text{max}]$. This is not required for ANODEC since its output remains within the feasible input range by design, see \eqref{eq:controller_ode}.

\subsection{Generalization to Tracking of Unseen Reference Signals}\label{sec:generalization_on_unseen_reference_signals}
A common criticism of data-driven or machine-learning-based solutions is that they might only perform well in the cases covered by the training dataset.
To verify that ANODEC generalizes beyond the step reference signals (used internally by ANODEC, see \eqref{eq:controller_learning}), we test the performance on two additional, qualitatively different reference signal distributions: Double-step and cubic-spline reference trajectories.
In total, 16 different reference signals are used to validate ANODEC.
One exemplary reference signal for each of the three reference signal distributions is illustrated in Fig.~\ref{fig:main_results} (dashed black line).

\begin{table}
\caption{Reference Tracking RMSE (in degrees) of PID and ANODEC for three reference signal distributions and two systems: Steps, double steps, and cubic splines. For each distribution, $N$ distinct reference signals are drawn and used to estimate the mean-RMSE and one standard deviation.}
\label{tab:res}
\begin{center}
\begin{sc}
\begin{tabular}{lccc}
\toprule
References & PID & ANODEC \\
\midrule
Setup 1 & & \\
Steps ($N=2$) & $12.89\pm6.14$ & $10.03\pm4.93$\\
Double Steps ($N=2$) & $12.86\pm2.66$& $11.08\pm1.37$\\
Cubic Splines ($N=12$) & $10.01\pm1.44$& \phantom{1}$4.48\pm1.11$\\
\midrule
Setup 2 & & \\
Steps ($N=2$)& $\phantom{0}4.00 \pm 0.45$ & $\phantom{0}5.54\pm0.53$\\
Double Steps ($N=2$) & $\phantom{0}9.50 \pm 4.03$ & $\phantom{0}6.81\pm0.75$\\
Cubic Splines ($N=4$) & $\phantom{0}4.56\pm0.81$ & $\phantom{0}5.01 \pm 0.79$\\
\bottomrule
\end{tabular}
\end{sc}
\end{center}
\end{table}

\subsection{ANODEC Reference Tracking Performance}\label{sec:sub_results}
We validate the proposed method~\cf{\ref{sec:anodec}} and compare to the baseline controller~\cf{\ref{sec:sub_pid_baseline}} for $16$ different reference signals~\cf{\ref{sec:generalization_on_unseen_reference_signals}} on the experimental setup of the PSA~\cf{\ref{sec:exp_setup}}.

ANODEC is able to outperform the PID controller baseline for all three reference signal distributions, as can be seen in Table~\ref{tab:res}.
Overall, ANODEC consistently outperforms the PID controller baseline in all 16 reference signals individually, and achieves a, on average, $\approx \num{45.9}\%$ lower RMSE tracking error across all reference signals while requiring less experimental interaction time (\qty{30}{\second} versus \qty{55}{\second}).
For each reference signal distribution, one exemplary reference signal is drawn. The performance of ANODEC compared to the PID controller baseline is shown in Figure~\ref{fig:main_results}.

\begin{figure}
    \centering
    \includegraphics[width=0.48\textwidth]{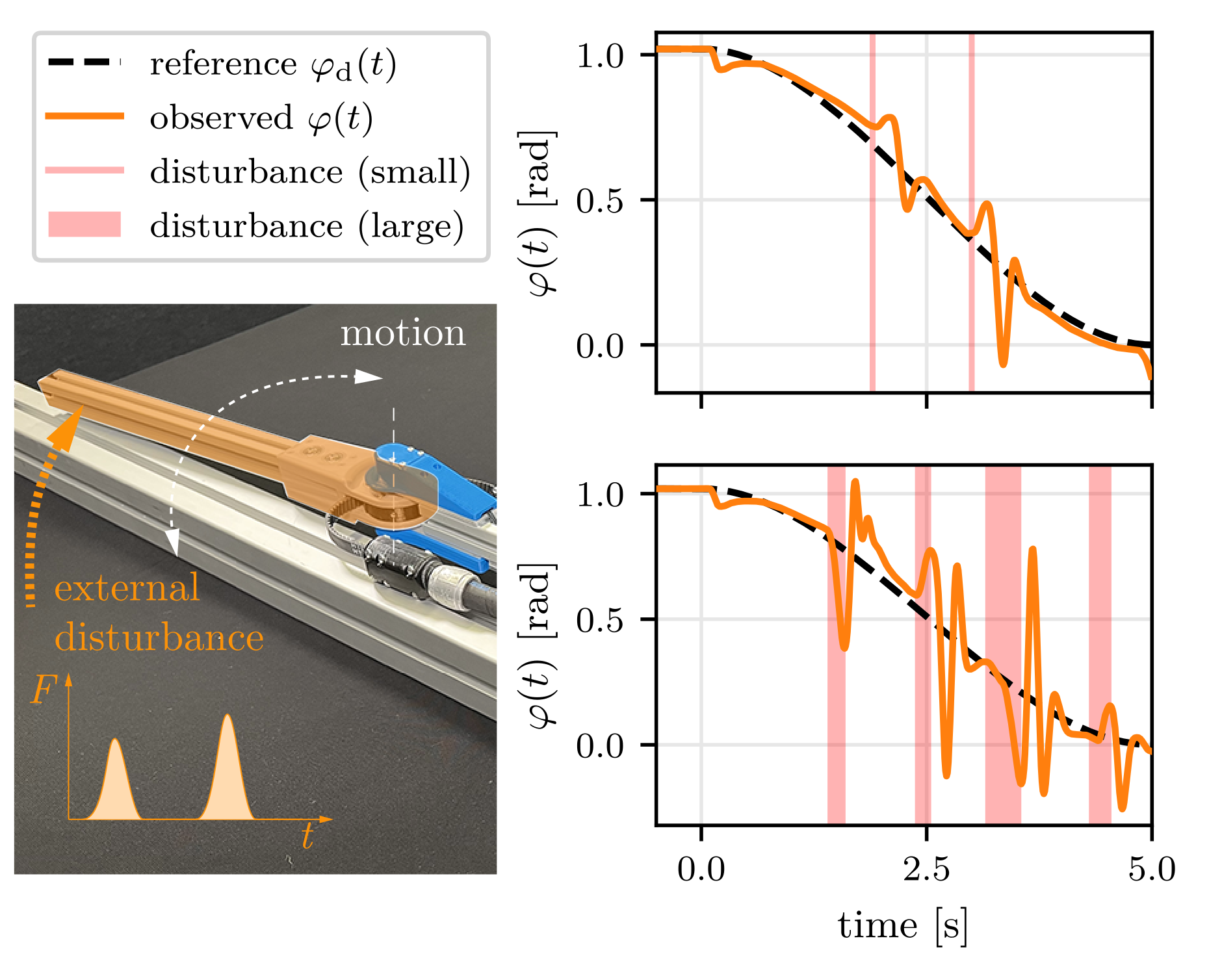}
    \vspace{-8mm}
    \caption{Tracking performance of ANODEC in two disturbed trials. In the first trial (top) there are two small disturbances (poking the endeffector with a stick) whereas in the second trial (bottom) there are four large disturbances (grabbing and briefly holding the endeffector). ANODEC remains stable and behaves as expected in both trials.}
    \label{fig:dist}
\end{figure}

Additionally, in Figure~\ref{fig:dist} we show that ANODEC is stable even if the trials are heavily disturbed.

These results show that ANODEC enables automatic yet competitive controller design for a PSA which, in contrast to conventional methods, requires no human expertise.

\subsection{System Variation}
To demonstrate the applicability of ANODEC to a second experimental system (Setup 2), we modify the experimental setup as it is described in Section~\ref{sec:exp_setup}. This variation significantly alters the system dynamics because the arm now rotates vertically, and the attached mass introduces a strong nonlinear term due to gravity.

The trial duration is extended from \qty{5}{\second} to \qty{8}{\second} and, as before, we record six input-output pairs (five for training, one for validation), now resulting in 48 seconds of data.
For this second use case, ANODEC successfully designs a competitive controller.
The achieved RMSEs are summarized in Table~\ref{tab:res}.

\section{Conclusion}
In this contribution, we have demonstrated that the data-driven method ANODEC can enable PSAs to -- fully automatically -- learn to perform agile, non-repetitive motions, from only \qty{30}{\second} of experimental interaction time.
Utilizing only input-output data and without any prior model knowledge, ANODEC designs a  single feedback controller that enables the tracking of arbitrary reference signals, which is then validated on multiple, qualitatively different, and even out-of-training-distribution reference signals.
Specifically, it is shown that ANODEC can outperform a manually tuned PID baseline and achieves an up to $\num{45.9}\%$ lower RMSE tracking error.
Overall, this contribution not only reinforces the validity of ANODEC but also represents a significant advancement towards developing practical, user-friendly PSAs capable of learning to perform agile movements with only a minimal amount of experimental interaction time.

Future work will involve validation on different experimental PSAs, including PSAs with multiple DOFs, and increasing the autonomy of ANODEC by, e.g., self-tuning of involved hyperparameters (such as $\lambda^{(c)}$) and autonomous data acquisition.

\printbibliography

\appendix

\begin{algorithm}
	\small
	\caption{Input Trajectory Generator}
	\label{alg:draw_u}
	\begin{algorithmic}
		\Require{$T \in \mathbb{R}_{> 0}$, $f \in \mathbb{N}_{> 0}$}
		\Function{generateSinusoidal}{}
		\Let{$\omega$} $\num{2} \pi f$
		\Let{us} $\sin(\omega \; \code{range}(0, T, 0.01)) \frac{\sqrt{\omega}}{2}$
		\State \Return{us}
		\EndFunction
    \State
		\Require{$T \in \mathbb{R}_{> 0}$, $\Delta t_\text{min} \in \mathbb{R}_{> 0}$, $\Delta t_\text{max} > \Delta t_\text{min}$, $u_{\text{min/max}} \in \mathbb{R}$}
		\Function{drawSpline}{}
		\Let{ts, us} $[0.0], [0.0]$
            \While{$\code{last}(\text{ts}) < T$}
                \Let{t} $\code{last}(\text{ts}) + \code{uniform}(\Delta t_\text{min}, \Delta t_\text{max})$
                \Let{u} $\code{coinflip}(\code{last}(\text{us}), \code{uniform}(u_\text{min}, u_\text{max}))$
                \Let{ts, us} $\code{append}(\text{ts}, \text{t})$, $\code{append}(\text{us}, \text{u})$
            \EndWhile
		\Let{us} $\code{cubicInterpolate}(\code{range}(0, T, 0.01), \text{ts}, \text{us})$
		\State \Return{us}
		\EndFunction
	\end{algorithmic}
\end{algorithm}

\end{document}